\UseRawInputEncoding
\RequirePackage{amsthm}
\documentclass[pdflatex,sn-mathphys-num]{sn-jnl}

\usepackage{graphicx}%
\usepackage{multirow}%
\usepackage{amsmath,amssymb,amsfonts}%
\usepackage{amsthm}%
\usepackage{mathrsfs}%
\usepackage[title]{appendix}%
\usepackage{xcolor}%
\usepackage{textcomp}%
\usepackage{manyfoot}%
\usepackage{booktabs}%
\usepackage{algorithm}%
\usepackage{algorithmicx}%
\usepackage{algpseudocode}%
\usepackage{listings}%
\usepackage{tikz}
\usepackage{xcolor}
\usepackage{threeparttable}
\usepackage{amsmath}
\usepackage{anyfontsize}
\usepackage{silence}
\theoremstyle{thmstyleone}%
\theoremstyle{thmstyletwo}%

\theoremstyle{thmstylethree}%

\begin{document}

\title[Article Title]{EchoMamba4Rec: Harmonizing Bidirectional State Space
Models with Spectral Filtering for Advanced Sequential
Recommendation}


\author[1]{\fnm{Yuda} \sur{Wang}}\email{q030026156@mail.uic.edu.cn}
\author[1]{\fnm{Xuxin} \sur{He}}\email{q030026051@mail.uic.edu.cn}
\author*[2,1]{\fnm{Shengxin} \sur{Zhu}}\email{Shengxin.Zhu@bnu.edu.cn}

\affil[1]{\orgname{Guangdong Provincial Key Laboratory of Interdisciplinary Research and Application for Data Science, BNU-HKBU United International College}, \orgaddress{\city{Zhuhai} \postcode{519087}, \country{China}}}
\affil[2]{\orgname{Research Center for Mathematics, Beijing Normal University}, \orgaddress{\street{No.18, Jingfeng Road}, \city{Zhuhai} \postcode{519087}, \state{Guangdong}, \country{China}}}

\abstract{Predicting user preferences and sequential dependencies based on historical behavior is the core goal of sequential recommendation. Although attention-based models have shown effectiveness in this field, they often struggle with inference inefficiency due to the quadratic computational complexity inherent in attention mechanisms, especially with long-range behavior sequences. Drawing inspiration from the recent advancements of state space models (SSMs) in control theory, which provide a robust framework for modeling and controlling dynamic systems, we introduce EchoMamba4Rec. Control theory emphasizes the use of SSMs for managing long-range dependencies and maintaining inferential efficiency through structured state matrices. EchoMamba4Rec leverages these control relationships in sequential recommendation and integrates bi-directional processing with frequency-domain filtering to capture complex patterns and dependencies in user interaction data more effectively. Our model benefits from the ability of state space models (SSMs) to learn and perform parallel computations, significantly enhancing computational efficiency and scalability. It features a bi-directional Mamba module that incorporates both forward and reverse Mamba components, leveraging information from both past and future interactions. 

\hspace{2em} Additionally, a filter layer operates in the frequency domain using Fast Fourier Transform (FFT) and learnable filters, followed by an inverse FFT to refine item embeddings and reduce noise. We also integrate Gate Linear Units (GLU) to dynamically control information flow, enhancing the model's expressiveness and training stability. Experimental results demonstrate that EchoMamba significantly outperforms existing models, providing more accurate and personalized recommendations. This study underscores the potential of integrating bi-directional and frequency-domain techniques with control theory to advance the state-of-the-art in sequential recommendation systems. Our code can be found at https://github.com/wyd0042/EchoMamba4Rec.}

\keywords{Sequential Recommendation, State Space Models, Bi-directional Processing, Frequency-domain Filtering}



\maketitle

\section{Introduction}\label{sec1}

\hspace{2em}Traditionally, research in recommendation systems has focused mainly on standard collaborative filtering, ranking methods \cite{chen2018survey} and regression methods \cite{gao2019learning, chen2020knowledge, chen2019censorious}. However, the landscape of recommendation systems has evolved significantly in recent years with the advent of deep learning and neural networks. Deep learning-based recommendation systems, such as XDeepFig \cite{xu2021xdeepfig} and deep interest models \cite{luo2023click, feng2022accelerating}, have shown better results by effectively capturing complex user-item interactions. Sequential recommendation is crucial in many real-world applications, such as e-commerce, social media \cite{zhang2021group}, medicine and drug system \cite{zheng2024},and content streaming platforms \cite{quadrana2018sequence}, where understanding user preferences and sequential dependencies can significantly enhance user experience and engagement. However, the sequential recommendation problem remains a relatively new yet important open research question, and it has proven challenging to leverage time and memory information to improve recommendation performance. Existing models, like those based on Transformers, face computational complexity constraints that limit their scalability and practicality, especially for long-range behavior sequences. Although Transformer models have achieved significant success in the recommendation systems domain \cite{liu2023linrec}, they face fundamental challenges when handling long sequences. In Transformers \cite{Vaswani2017}, each token can refer to all preceding tokens during prediction, leading to a quadratic time complexity of O($n^2$) during training, often called the ``quadratic bottleneck'' \cite{keles2023computational}. Additionally, the key-value (KV) cache calculation in attention models required to store these tokens incurs an O(n) space complexity. This results in increased memory usage, raising the risk of CUDA Out-of-Memory (OOM) errors as memory consumption grows. Thus, exploring efficient and effective sequential recommendation solutions is highly relevant and valuable.

 Addressing this performance-efficiency impasse, the state space model (SSM) \cite{hamilton1994state,gu2021efficiently} emerges as a central operator, heralding an era of efficient sequential recommendation. Originating from control theory \cite{williams2007linear}, SSMs are designed to model dynamic systems by representing them with state variables and equations that describe their evolution over time \cite{GuDao2023}. This approach offers significant advantages in maintaining inferential efficiency and managing long-range dependencies via structured state matrices. With a lineage tracing back to alternatives for RNNs and Transformers across various language tasks, the allure of SSM lies in its ability to encapsulate system dynamics accurately while ensuring computational feasibility.This paper is the vanguard in harnessing selective SSMs, particularly Mamba \cite{GuDao2023}—a variant with data-dependent selectivity and efficient hardware-aligned algorithms—for the purpose of sequential recommendation \cite{Liu2024}. The convergence of the Mamba block with novel sequential modeling techniques illustrates a significant leap forward, enhancing the model's performance without compromising on inferential swiftness. The research delineates the overarching contributions of our EchoMamba4Rec model, demonstrates its supremacy over RNN and attention-based baselines through experimental analysis, and sets the stage for advancing SSM-based models in the recommender system sphere.

 The Mamba model employs a state space model (SSM) inspired by control theory to replace the traditional attention mechanism, while retaining multi-layer perceptron (MLP) style projections for computation. This approach allows Mamba to efficiently handle sequential data. The Mamba model has the following three key highlights \cite{GuDao2023}:

\begin{enumerate}
  \item \textbf{Long Sequence Processing}: Mamba can process token sequences as long as millions of tokens, which is critical for tasks that require long-term memory.

  \item \textbf{Fast Inference}: Mamba's inference speed is exceptionally fast, about five times faster than that of Transformer models, which means it can process more data in a shorter time \cite{zhou2021informer, GuDao2023}.

  \item \textbf{Linear Scalability}: Mamba has linear scalability with respect to sequence length \cite{GuDao2023}, indicating that as the sequence length increases, the model's performance does not significantly degrade.
\end{enumerate}

The main contributions of this paper are as follows:
We enhance existing models that utilize Mamba for sequential recommendation by incorporating a Fourier layer and Gate Linear Units (GLU), and by employing bi-directional processing strategies. These improvements bolster the sequence modeling capability while maintaining inference efficiency and address the prevalent effectiveness-efficiency dilemma existing in RNNs and Transformers.

\section{Related Work}\label{sec2}

\subsection{Sequential Recommander with deep learning}
The evolution of recommendation systems has been significantly influenced by advancements in deep learning, marking a shift towards more personalized and dynamic user experiences. Early neural approaches to sequential recommendation utilized Convolutional Neural Networks (CNNs) and Recurrent Neural Networks (RNNs) to model user behaviors, paving the way for the application of neural networks in this domain. However, these models often grappled with challenges such as catastrophic forgetting and limited ability to handle long-term dependencies \cite{Hidasi2015, Kirkpatrick2017}. The introduction of transformer models, with their self-attention mechanisms, represented a leap forward, offering remarkable performance improvements by effectively capturing dynamic user interactions \cite{Vaswani2017, Kang2018}. CycleTrans has proven successful in dealing with sequential Electronic Health Records (EHR) \cite{zheng2024}, showcasing the potential of transformer-based models in sequential recommendation tasks. Despite their success, attention-based methods encounter inference inefficiency due to the quadratic computational complexity inherent in attention operators, particularly for long user behavior sequences or large-scale datasets \cite{Liu2024}.

\subsection{State Space Model}
State Space Models (SSM) provide a compelling framework for modeling sequences, offering significant advantages over transformer models due to their linear computational complexity \cite{hamilton1994state}. Unlike transformers, which exhibit quadratic complexity $O(n^2 \cdot d)$ with respect to sequence length due to their self-attention mechanism, SSM operates with a complexity of $O(n \cdot d^2)$, making them ideal for long sequences.

 In control theory, SSMs are pivotal for representing dynamic systems through state variables, which encapsulate the system's status at any given time. This representation comprises a set of first-order differential (or difference) equations that describe the evolution of the state vector and an output equation that maps the state vector to observed data 
 \cite{kalman1960new}. These models enable efficient prediction and control of system behavior over time, making them ideal for applications in signal processing, econometrics, and time series analysis 
 \cite{ogata2002modern}. The ability to incorporate noise and other disturbances into the model further enhances their applicability in real-world scenarios 
 \cite{ljung2010perspectives}.

 Furthermore, linear state-space control systems are fundamental for the analysis and design of linear control systems, making them ideal for both researchers and students preparing for advanced studies in systems and control theory \cite{williams2007linear}.
The foundational continuous form of SSM is defined by the linear ordinary differential equations:
\begin{align}
h'(t) &= Ah(t) + Bx(t), \\
y(t) &= Ch(t),
\end{align}
where $h(t)$ represents the latent state, $x(t)$ the input sequence, and $y(t)$ the output, with $A$, $B$, and $C$ as learnable matrices.

 To adapt this model for digital computation, it is discretized using a step size $\Delta$, resulting in:
\begin{align}
h_t &= Ah_{t-1} + Bx_t, \\
y_t &= Ch_t,
\end{align}
where $A$ and $B$ are modified to $A = \exp(\Delta A)$ and $B = (\Delta A)^{-1} (\exp(\Delta A) - I) \Delta B$.

 Enhancements such as the Structured State Space Model (S4) \cite{gu2021efficiently} and Mamba further optimize SSM. S4 introduces a structured approach to the state transition matrix $A$ using HiPPO techniques \cite{gu2021efficiently,gu2020hippo}, improving long-range dependency modeling. Mamba extends S4 by incorporating a data-dependent selection mechanism and a hardware-aware parallel algorithm, maintaining linear complexity while enhancing data handling efficiency.

 These developments position SSMs as superior for tasks involving long sequences where computational efficiency is paramount, offering Transformer-level performance with greater scalability and lower resource consumption.

\subsection{Mamba Block}
The Mamba block emerges as a novel solution to the aforementioned challenges of catastrophic forgetting, limited ability to handle long-term dependencies, and inference inefficiency in long user behavior sequences or large-scale datasets, leveraging the capabilities of state space models (SSMs) for efficient sequential processing. Originating from the structured state space model (S4) \cite{gu2021efficiently}, the Mamba block introduces a data-dependent selective mechanism, enabling the model to adaptively focus on relevant information while filtering out noise \cite{GuGoel2021}. This selective SSM, combined with a hardware-aware parallel algorithm, allows for significant improvements in both computational efficiency and model performance, as shown in Figure \ref{mamba}. The Mamba block's efficiency is further enhanced by its linear-time complexity in sequence modeling, offering a compelling alternative to the more computationally demanding Transformer and RNN architectures. By incorporating structured state matrices initialized with HiPPO for improved long-range dependency modeling, the Mamba block demonstrates Transformer-quality performance with greater efficiency, particularly in processing long sequences \cite{Liu2024}.

\begin{figure}[h]
\centering
\begin{tikzpicture}

  \draw[rounded corners, fill=gray!10] (5.75,0) rectangle (9,5);
    \filldraw[fill=green!20, draw=black] 
        (6.25,0.25) -- (7,0.25) -- (7.25,0.75) -- (6,0.75) -- cycle;
    \node at (6.625, 0.5) {LP};

    \filldraw[fill=green!20, draw=black] 
        (7.75,0.25) -- (8.5,0.25) -- (8.75,0.75) -- (7.5,0.75) -- cycle;
    \node at (8.125, 0.5) {LP};
    \node[draw, fill=blue!20, minimum width=1.5cm, minimum height=0.4cm] (con) at (6.625,1.4) {Conv};
    \node (silu1) [circle, draw, fill=white!30, minimum width=0.5cm, minimum height=0.5cm] at (6.625, 2.2) {$\sigma$};
    \node (silu2) [circle, draw, fill=white!30, minimum width=0.5cm, minimum height=0.5cm] at (8.125, 2.2) {$\sigma$};
    \node[draw, fill=blue!20, minimum width=1.5cm, minimum height=0.4cm] (ssm) at (6.625,3) {SSM};
    \node (silu3) [circle, draw, fill=white!30, minimum width=0.5cm, minimum height=0.5cm] at (6.625, 3.7){ };
    \draw[thick] (silu3.north west) -- (silu3.south east);
    \draw[thick] (silu3.north east) -- (silu3.south west);
    \filldraw [fill=green!20, draw=black] 
        (6.25,4.75) -- (7,4.75) -- (7.25,4.25) -- (6,4.25) -- cycle;
    \node at (6.625, 4.5) {LP};

    \draw[-] (con) -- (silu1);
    \draw[-] (silu1) -- (ssm);
    \draw[-] (ssm) -- (silu3);
    \draw[-] (silu3) -- (6.625, 4.25);
    \draw[-] (6.625, 0.75) -- (con);
    \draw[-] (8.125, 0.75) -- (silu2);
    \draw[-] (silu2) |- (silu3);
    \draw[-] (6.625, -0.25) -- (6.625, 0.25);
    \draw[-] (8.125, 0.25) |- (6.625, 0.12);
    \draw[->] (6.625, 4.75) -- (6.625, 5.25);

    \node (silu4) [circle, draw, fill=white!30, minimum width=0.1cm, minimum height=0.1cm] at (6.2, 5.6) {$\sigma$};
    \node at (7.8,5.6) {\small SiLU activation};
    \node (silu5) [circle, draw, fill=white!30, minimum width=0.5cm, minimum height=0.5cm] at (6.2, 6.3){ };
    \draw[thick] (silu5.north west) -- (silu5.south east);
    \draw[thick] (silu5.north east) -- (silu5.south west);
   \node at (7.8,6.3) {\small Nonlinearity}; 
   \node at (8.9,-0.75) {\shortstack{\small Linear\\ projection}}; 
   \filldraw [fill=green!20, draw=black] 
    (5.625,-0.5) -- (6.375,-0.5) -- (6.625,-1) -- (5.375,-1) -- cycle; 
   \node at (6.0, -0.75) {LP}; 
   \filldraw[fill=green!20, draw=black] 
    (6.925,-1.0) -- (7.675,-1.0) -- (7.925,-0.5) -- (6.675,-0.5) -- cycle; 
   \node at (7.3, -0.75) {LP};

\end{tikzpicture}
\centering
\caption{\textbf{Mamba block} \cite{GuDao2023}}
\label{mamba}
\end{figure}

\section{Methodology}\label{sec3}

In this section, we showcase our proposed framework, EchoMamba4Rec. We start with a high-level overview of the framework, followed by a detailed examination of its technical components. We explain how EchoMamba4Rec builds a sequential recommendation model using an embedding layer, selective state space models, and a prediction layer. Additionally, we cover key components commonly used in sequential recommendation, such as positional embeddings, feed-forward networks, dropout, and layer normalization.
\subsection{Model definition}
Consider a user set \( U = \{u_1, u_2, \ldots, u_{|U|}\} \) and an item set \( V = \{v_1, v_2, \ldots, v_{|V|}\} \). For each user \( u \in U \), their interaction sequence in chronological order is denoted as \( S_u = [v_1, v_2, \ldots, v_{n_u}] \), where \( n_u \) is the length of the sequence. Given the interaction history \( S_u \), the standard task is to predict the next item \( v_{n_u+1} \) that user \( u \) will interact with.
\subsection{Framework Overview}

\begin{figure*}[h]
\centering
\resizebox{\textwidth}{!}{%
\begin{tikzpicture}
 \definecolor{mygold}{rgb}{1.0, 0.84, 0.0}
  \definecolor{mypurple}{rgb}{0.5, 0.0, 0.5}
  \draw[rounded corners, fill=blue!20] (1,-2) rectangle (4,-1.5);
  \node at (2.5,-1.75) {Embedding Layer};

  \draw[rounded corners, fill=blue!20] (1,6.5) rectangle (4,7);
  \node at (2.5,6.75) {Prediction Layer};

  \draw[rounded corners, fill=gray!10] (-0.4,-0.8) rectangle (5,6.2);
  \draw[rounded corners, fill=gray!10] (-0.2,-1) rectangle (5.2,6);
  
  \node at (5,6.4) {x L};

  \node[draw, rounded corners, fill=mygold!50, minimum width=3cm, minimum height=0.4cm] (fl) at (2.5,-0.35) {Filter Layer};

  \node[draw, rounded corners, fill=yellow!30, minimum width=1.5cm, minimum height=0.4cm] (addnorm1) at (2.5,5.5) {Add \& Norm};
  \node[draw, rounded corners, fill=purple!30, minimum width=1.5cm, minimum height=0.4cm] (ffn) at (2.5,4.5) {GLU};

  \node[draw, rounded corners, fill=yellow!30, minimum width=1.5cm, minimum height=0.4cm] (addnorm2) at (1.2,2.9) {Add \& Norm};
  \node[draw, rounded corners, fill=yellow!30, minimum width=1.5cm, minimum height=0.4cm] (addnorm3) at (3.8,2.9) {Add \& Norm};

  \node[draw, rounded corners, fill=pink!80, minimum width=1.5cm, minimum height=0.4cm] (mamba1) at (1.2,1.6) {Mamba};
  \node[draw, rounded corners, fill=pink!80, minimum width=1.5cm, minimum height=0.4cm] (mamba2) at (3.8,1.6) {Mamba};  

  \node[draw, circle, fill=white, minimum size=0.4cm] (circle) at (2.5,3.67){  };

  \draw[->] (3.8, 0.82) -- (mamba2);%
  \draw[->] (mamba1) -- (addnorm2);
  \draw[->] (ffn) -- (addnorm1);
  \draw[->] (1.2,0.82) -| (mamba1);%
  \draw[-] (fl) -- (2.5,0.35); 
  \draw[->] (2.5,5.72) -- (2.5,6.5); 
  \draw[->] (addnorm2) |- (circle);
  \draw[->] (addnorm3) |- (circle);
  \draw[->] (2.5,3.46) -- (2.5,4.28); 
  \draw[-] (2.2,3.67) -- (2.8,3.67);
  \draw[dashed, ->] (2.5,0.35) |- (addnorm2);%
  \draw[dashed, ->] (2.5,0.35) |- (addnorm3);%
  \draw[dashed, -] (2.5,4) -- (3.8,4);
  \draw[dashed, ->] (3.8,4) |- (addnorm1);
  \draw[->] (2.5,-1.5) -- (fl);
  \draw[->] (2.5,-2.3) -- (2.5,-2);
  \draw[->] (2.5,7) -- (2.5,7.3);
  \draw[-] (2.5,0.35) -| (1.2, 0.62);%
  \draw[-] (2.5,0.35) -| (3.8, 0.62);%
  \draw[->] (mamba2) -- (addnorm3);%

  \node at (1.2, 0.75) {\small{\( v_1, v_2, \ldots, v_{n_u} \)}};
  \node at (3.85, 0.75) {\small{\( v_{n_u}, \ldots, v_2, v_1\)}};
  \node at (4.5, 2.2) {\textbf \small{Reverse}};
  \node at (4.5, 0.35) {\textbf \small{Reverse}};
  
  \draw[rounded corners, fill=gray!10] (5.75,2.7) rectangle (9.4,5.65);

  \node[draw, rounded corners, fill=orange!30, minimum width=1cm, minimum height=0.3cm] (con1) at (7.6,3.96) {\scriptsize A=${Y}_t^{final}{W}_1$+${b}_1$};
  \node[draw, rounded corners, fill=purple!30, minimum width=2cm, minimum height=0.1cm] (con) at (7.6,3.3) {\small {B=${Y}_t^{final} {W}_2$+${b}_2$}};

  \node (silu2) [circle, draw, fill=white!30, minimum width=0.5cm, minimum height=0.5cm] at (8.6, 5.2) {$\sigma$};
  \node (silu3) [circle, draw, fill=white!30, minimum width=0.5cm, minimum height=0.5cm] at (7.6, 5.2) { };
  \draw[thick] (silu3.north west) -- (silu3.south east);
  \draw[thick] (silu3.north east) -- (silu3.south west);

  \draw[->] (con1) -- (silu3);

  \draw[->] (5.95, 2.5) |- (6.26, 3.63);
  \draw[-] (con) -- (9.2,3.3);
  \draw[-] (9.2,3.3) |- (silu2);
  \draw[->] (silu2) -- (silu3);

  \draw[->] (silu3) -- (7.6, 6);

  \node (silu4) [circle, draw, fill=white!30, minimum width=0.1cm, minimum height=0.1cm] at (6.2, 2) {$\sigma$};
  \node at (8,2) {\small Sigmoid function};
  \node (silu5) [circle, draw, fill=white!30, minimum width=0.5cm, minimum height=0.5cm] at (6.2, 1.3) { };
  \draw[thick] (silu5.north west) -- (silu5.south east);
  \draw[thick] (silu5.north east) -- (silu5.south west);
  \node at (8,1.3) {\small \shortstack{Element-wise\\ multiplication}};

  \draw[dashed, -] (5.75,2.7) -- (3.25, 4.3);
  \draw[dashed, -] (5.75,5.65) -- (3.25, 4.7);

  \draw[rounded corners, fill=gray!10] (-4,-1.5) rectangle (-1,4.5);

  \node[draw, rounded corners, fill=purple!30, minimum width=2cm, minimum height=0.4cm] (Layer Norm) at (-2.5,4) {GLU};
  \node[draw, rounded corners, fill=yellow!30, minimum width=2cm, minimum height=0.4cm] (GLU) at (-2.5,3) {Add \& Norm};

  \node[draw, rounded corners, fill=mygold!50, minimum width=2cm, minimum height=0.4cm] (iFFT) at (-2.5,1.8) {Inverse FFT};
  \node[draw, rounded corners, fill=mypurple!30, minimum width=2.5cm, minimum height=0.4cm] (Multiple) at (-2.5,0.35) {\shortstack{Multiple\\Learnable Filter}};
  \node[draw, rounded corners, fill=mygold!50, minimum width=2cm, minimum height=0.4cm] (FFT) at (-2.5,-1) {FFT};

  \draw[->] (FFT) -- (Multiple);
  \draw[->] (Multiple) -- (iFFT);
  \draw[->] (iFFT) -- (GLU);
  \draw[->] (GLU) -- (Layer Norm);
  \draw[->] (-2.5, -1.7) -- (FFT);
  \draw[-] (-2.5, -1.7) -- (-4.2, -1.7);
  \draw[->] (-4.2, -1.7) |- (GLU);
  \draw[->] (Layer Norm) |- (-2.5, 4.9);
  \draw[dashed, -] (-1,4.5) -- (1.1,-0.1);
  \draw[dashed, -] (-1,-1.5) -- (1.1,-0.5);

  \node at (-2.1,2.4) {\small $\tilde{F}^{l}$}; 
  \node at (-2.1,1.2) {\small $\tilde{X}^{l}$}; 
  \node at (-2.1,-0.4) {\small $X^{l}$};

\end{tikzpicture}}
\centering
\caption{\textbf{EchoMamba4Rec}.  The process starts with embedding user information using an recommendation embedding layer. Next, a filter layer was utilized to extracts essential sequence information. This filtered data is then processed by the bi-directional EchoMamba block, handling sequences in both forward and reverse directions. A Gated Linear Unit (GLU) is used to dynamically control information flow, enhancing the model's expressiveness and stability. The final step involves combining and normalizing the processed data before generating the sequential recommendation output. Compared to Mamba4Rec, our model places greater emphasis on extracting sequence features while reducing noise, thereby improving model accuracy and robustness.}
\label{f1}
\end{figure*}
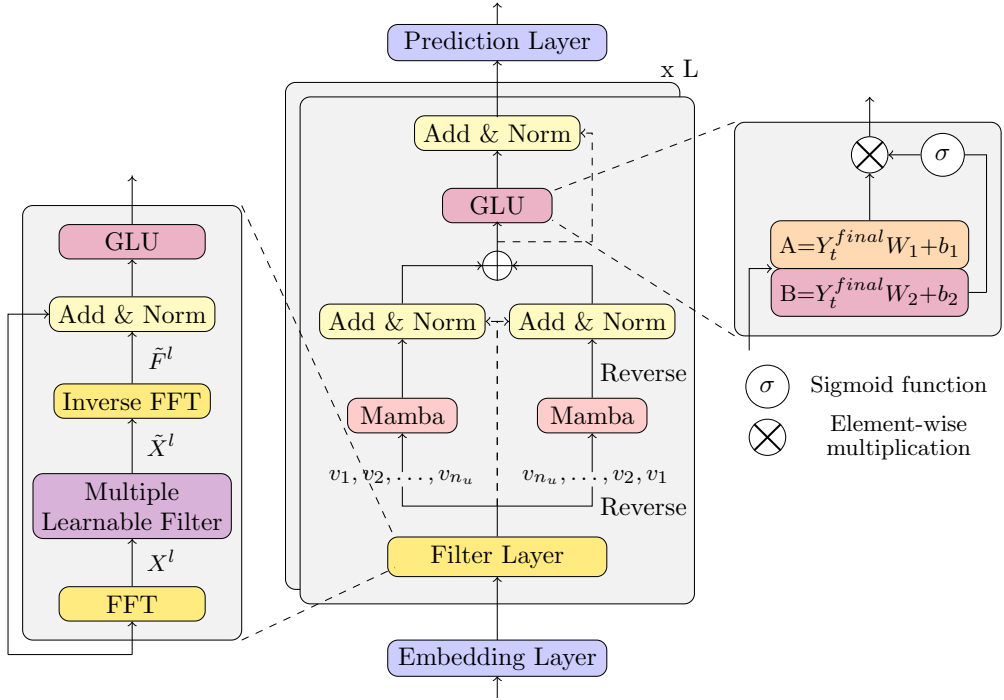

As illustrated in the Figure \ref{f1}, the proposed EchoMamba4Rec is a sequential recommendation model that leverages selective state space models through Mamba blocks. The core element of EchoMamba4Rec is the Mamba layer, which combines a Mamba block with a position-wise feed-forward network. EchoMamba4Rec offers flexibility in its bidirectional architecture: it can be constructed by flexibly combining different types of recurrent neural network (RNN), long short-term memory network (LSTM) \cite{hochreiter1997long}, gated recurrent unit (GRU) \cite{chung2014empirical} and other structures according to task requirements and data characteristics to adapt to different scenarios and data characteristics. Each layer comprises a Mamba block followed by a gate linear unit (GLU), enabling the model to capture both item-specific information and sequential context effectively from the user’s interaction history.

\subsection{Embedding Layer}

Our method employs an embedding layer to project item IDs into a high-dimensional space. This embedding layer utilizes a learnable embedding matrix \(\mathbf{E} \in \mathbb{R}^{|V| \times D}\), where \(D\) denotes the embedding dimension. By applying the embedding layer to the input item sequence \(S_u\), we obtain the initial item embeddings \(\mathbf{H}_o \in \mathbb{R}^{n_u \times D}\), where \(n_u\) is the number of items in the sequence for user \(u\).

To improve robustness and prevent overfitting, we apply dropout to these embeddings. Dropout is a regularization technique that randomly sets a fraction of the input units to zero at each update during training, which helps to prevent the model from becoming too sensitive to the specific weights of neurons. The dropout rate determines the fraction of neurons to drop, typically set between 0.2 and 0.5.

After applying dropout, we normalize the embeddings using layer normalization. Layer normalization standardizes the activations of the previous layer for each individual data point, which helps stabilize and accelerate training by reducing internal covariate shift. The resulting embeddings \(\mathbf{H}\) are computed as follows:

\[
\mathbf{H} = \text{LayerNorm}(\text{Dropout}(\mathbf{H}_o)) \in \mathbb{R}^{n_u \times D}. \tag{5}
\]

During both training and inference phases, we aggregate multiple samples into a mini-batch, resulting in input data \(\mathbf{X} \in \mathbb{R}^{B \times L \times D}\) for the subsequent Mamba block, where \(B\) represents the batch size and \(L\) is the padded sequence length.

\subsection{Bi-directional Modeling}
In sequential recommendation systems, it is crucial to capture dependencies and patterns not only from past interactions but also considering future context. That's what ``Echo'' comes out. A bi-directional model addresses this by processing sequences from both directions: forward and reverse.

\subsubsection{Reverse Sequence Modeling}
To enhance the prediction capability, we propose employing a reverse sequence modeling approach where the interaction sequence for user \( u \) is processed in reverse order, denoted as \( S_u^R = [v_{n_u}, v_{n_u-1}, \ldots, v_1] \). The reverse model predicts the same next item \( v_{n_u+1} \), considering the sequence from the most recent interaction back to the first. This reverse perspective allows the model to capture different dependencies that may be overlooked when only considering the forward sequence.

\paragraph{Rationale for Bi-directional Approach}
The rationale behind using a bi-directional model is twofold:

\noindent \textbf{Enhanced Contextual Understanding:} by analyzing user behavior from both the start and end of the sequence, the model gains a comprehensive view, capturing patterns that emerge over time and those that are most immediate.

\noindent \textbf{Robustness to Sequence Variability:} some user behaviors are better understood when the context leading up to and following certain actions is considered. Bi-directional models are more robust to changes in user behavior patterns, providing stability in learning and prediction.

 The combination of forward and reverse sequence modeling in a bi-directional framework allows for a more nuanced understanding of user interactions, potentially improving the accuracy of the predictions in dynamic recommendation environments.

\subsection{Filter Layer}
The second reason we named our model EchoMamba4Rec is due to our use of a Fourier Transform-based (FFT) filter \cite{zhou2022filter} to capture sequential dependencies. In the filter layer, a filtering operation is performed for each feature dimension in the frequency domain, followed by the application of skip connections and layer normalization. Given the input item representation matrix $\mathbf{F}^l \in \mathbb{R}^{n \times d}$ at layer $l$ (where $l=0$ implies $\mathbf{F}^0 = \mathbf{E_I}$), where $n$ is the number of items and $d$ is the feature dimension, we first perform a one-dimensional FFT along the item dimension to convert $\mathbf{F}^l$ into the frequency domain:

\[
\mathbf{X}^l = \mathcal{F}(\mathbf{F}^l) \in \mathbb{C}^{n \times d}, \tag{6}
\]

\noindent where $\mathcal{F}(\cdot)$ denotes the one-dimensional FFT, and $\mathbf{X}^l$ is a complex tensor representing the spectrum of $\mathbf{F}^l$. We then modulate this spectrum by multiplying it with a filter $\mathbf{K} \in \mathbb{C}^{n \times d}$ which is learnable:

\[
\mathbf{\tilde{X}}^l = \mathbf{K} \odot \mathbf{X}^l, \tag{7}
\]

\noindent where $\odot$ represents element-wise multiplication. The modulated spectrum $\mathbf{\tilde{X}}^l$ is transformed back to the time domain using the inverse FFT to update the sequence representations:

\[
\mathbf{\tilde{F}}^l \leftarrow \mathcal{F}^{-1}(\mathbf{\tilde{X}}^l) \in \mathbb{R}^{n \times d}, \tag{8}
\]

\noindent $\mathcal{F}^{-1}(\cdot)$ denotes the inverse 1D FFT, converting the complex tensor back into a real number tensor. The operations of FFT and inverse FFT help reduce noise from the logged data, enabling cleaner item embeddings. We incorporate skip connections, layer normalization, and dropout operations to mitigate problems related to gradient vanishing and unstable training:

\[
\mathbf{\tilde{F}}^l = \text{LayerNorm}(\mathbf{F}^l + \text{Dropout}(\mathbf{\tilde{F}}^l)), \tag{9}
\]
\subsection{Gate Linear Unit}
\noindent After the previous section, we process the data through a GLU network, which has a gating mechanism that can selectively retain or discard some parts of the input data, thus effectively suppressing the overfitting phenomenon of the model:

\[
\text{GLU}(\mathbf{\tilde{F}}^l) = (\mathbf{\tilde{F}}^l \mathbf{W}_1 + \mathbf{b}_1) \odot \sigma(\mathbf{\tilde{F}}^l \mathbf{W}_2 + \mathbf{b}_2), \tag{10}
\]

\noindent where $\odot$ represents element-wise multiplication, $\sigma$ is the sigmoid function, $\mathbf{W}_1$ and $\mathbf{W}_2$ are learnable weight matrices, and $\mathbf{b}_1$ and $\mathbf{b}_2$ are bias terms. $\mathbf{\tilde{F}}^l$ represents the input feature map from the previous layer.

\noindent Note: \( \tilde{F} \) and \( \tilde{X} \) indicate the filtered or modulated versions of the original representations after the filtering operation.


\subsection{EchoMamba Block}

The EchoMamba block operates on an input $\mathbf{F}^l \in \mathbb{R}^{B \times L \times D}$, where $B$ represents the batch size, $L$ denotes the sequence length, and $D$ signifies the hidden dimension at each time step $t$. 

\[
\mathbf{X}' = \text{Conv1D}(\mathbf{F}^l), \tag{11}
\]

 It begins by performing linear projections on the input $\mathbf{F}^l$ with an expanded hidden dimension to obtain $\mathbf{X}'$. This projection is then processed through a 1D convolution and a SiLU activation. The SiLU (Sigmoid Linear Unit) activation function is defined as:

\[
\text{SiLU}(\mathbf{X}') = \mathbf{X}' \cdot \sigma(\mathbf{X}'), \quad \text{where} \quad \sigma(\mathbf{X}') = \frac{1}{1 + e^{-\mathbf{X}'}},
\tag{12}
\]

The core of the block involves a selective state space model (SSM) with parameters discretized based on the input. 

\[
\mathbf{H} = \text{SSM}(\mathbf{X}'), \tag{13}
\]

 This discretized SSM generates the state representation $\mathbf{H}$. Finally, $\mathbf{H}$ is combined with a residual connection from $\mathbf{X}'$ after applying SiLU, and a final linear projection delivers the final output $\mathbf{Y}_t$ at time step $t$.

\[
\mathbf{Y}_t = \text{Linear}(\text{SiLU}(\mathbf{H}) + \mathbf{X}'), \tag{14}
\]

To further enhance the model, the EchoMamba block employs bidirectional processing. This means it processes sequences in both forward and reverse order. For a given user sequence $u = [v_{n_u}, v_{n_u-1}, \ldots, v_1]$, the model captures dependencies from both directions. The forward sequence is processed as described above. For the reverse sequence, we define:

\[
\mathbf{X}'_{rev} = \text{Conv1D}(\mathbf{F}^l_{rev}), \tag{15}
\]

\[
\mathbf{H}_{rev} = \text{SSM}(\mathbf{X}'_{rev}), \tag{16}
\]

\[
\mathbf{Y}_{t, rev} = \text{Linear}(\text{SiLU}(\mathbf{H}_{rev}) + \mathbf{X}'_{rev}). \tag{17}
\]

\noindent The final output combines both forward and reverse sequences:

\[
\mathbf{Y}_t^{final} = \text{Linear}(\text{Concat}[\mathbf{Y}_t, \mathbf{Y}_{t, rev}]). \tag{18}
\]

\noindent The final output then passes through a GLU to further refine the features:

\[
\text{GLU}(\mathbf{Y}_t^{final}) = (\mathbf{Y}_t^{final} \mathbf{W}_1 + \mathbf{b}_1) \odot \sigma(\mathbf{Y}_t^{final} \mathbf{W}_2 + \mathbf{b}_2). \tag{19}
\]

 Overall, the EchoMamba block leverages input-dependent adaptations and a selective SSM to process sequential information effectively. The parameters of the EchoMamba block consist of an SSM state expansion factor $N$, a kernel size $K$ for the convolution, and a block expansion factor $E$ for input and output linear projections. Selecting the expansion factors $N$ and $E$ involves a tradeoff between capturing complex relationships and training efficiency, where higher $N$ and $E$ increase computational cost.

 The EchoMamba block addresses the prevalent effectiveness-efficiency dilemma existing in RNNs and Transformers. The recurrent SSM in the EchoMamba block operates in linear time with respect to the sequence length, making it as efficient as RNNs for inference, achieved by discretized parameters. Additionally, the selection mechanism makes the parameters of SSM data-dependent. This enables EchoMamba4Rec to selectively retain or discard information, which is crucial for capturing the complex dependencies inherent in sequential recommendation tasks. By leveraging the structured state matrix with HiPPO techniques, EchoMamba block effectively models long-range dependencies within user interaction sequences, surpassing the performance of traditional self-attention mechanisms in handling long sequences \cite{Liu2024}.

\subsection{Prediction Layer}

In the final layer of EchoMamba4Rec, we compute the user's preference score for the next item \( v_{n_u+1} \) given their interaction history. The computation leverages both the bi-directional processing and the filtered embeddings obtained from previous layers.

\noindent Given the user \( u \)'s interaction sequence \( S_u = [v_1, v_2, \ldots, v_{n_u}] \) and the bi-directionally processed representations, the prediction is formulated as follows:

\[
P(v_{n_u+1} = v | v_1, v_2, \ldots, v_{n_u}) = \mathbf{e}_v^\top \mathbf{Y}_t^{final}, \tag{20}
\]

\noindent where \( \mathbf{e}_v \) represents the embedding of item \( v \) from the item embedding matrix \( \mathbf{M}_I \), and \( \mathbf{Y}_t^{final} \) is the final output of the GLU-enhanced bi-directional blocks for user \( u \).

\noindent The final output $\mathbf{Y}_t^{final}$ combines both the forward and reverse bi-directional processing results, providing a comprehensive representation of the user's interaction history. This approach allows the model to effectively capture complex dependencies and improve the accuracy of sequential recommendations.




\section{Experiments}\label{sec4}

\subsection{Experimental Setup}
\textbf{Datasets.}  Three datasets contain product reviews and ratings for the “Beauty”, “Video Games” categories on Amazon (\textbf{Amazon-Beauty} and \textbf{Amazon-Video-Games}) \cite{ni2019justifying} and movie ratings from MovieLens (\textbf{MovieLens-1M}) \cite{umemoto2022ml}, which analyzed using the RecBole framework \cite{zhao2021recbole}. For each user, the interaction sequence is constructed by sorting their interaction records according to their timestamps. Users and items with less than 5 interactions are filtered. The preprocessed statistics of each dataset are summarized in Table \ref{tab:Statistics}.

\vspace{0.3cm} %

\begin{table}[h!]
    \centering
    \caption{Statistics of the Amazon Review Datasets for Beauty and Video Games categories and MovieLens Datasets for movie ratings. These datasets provide comprehensive user-item interaction data, facilitating research on sequential recommendation systems. The statistics include the number of users, items, interactions, and the average interaction length per user.}
    \begin{tabular}{lcccc}
        \toprule
        \textbf{Dataset} & \textbf{\# users} & \textbf{\# items} & \textbf{\# interactions} & \textbf{Avg. Length} \\
        \midrule
        MovieLens-1M & 6,040 & 3,416 & 999,611 & 165.4 \\
        Amazon-Beauty & 22,363 & 12,101 & 198,502 & 8.9 \\
        Amazon-Video-Games & 14,494 & 6,950 & 132,209 & 9.1 \\
        \bottomrule
    \end{tabular}
    \label{tab:Statistics}
\end{table}

\noindent\textbf{Hyperparameters.} The Adam optimizer is used with a learning rate 0.001. The training batch size is 2048, and the evaluation batch size is 4096. All models use an embedding dimension of 64. And the dropout rate is 0.2.

\vspace{0.3cm} %

\noindent\textbf{Baselines.} We compare EchoMamba4Rec against several baseline methods, which include RNN-based models like GRU4Rec \cite{jannach2017recurrent} and NARM \cite{li2017neural} , as well as attention-based models such as SASRec \cite{Kang2018} and BERT4Rec \cite{sun2019bert4rec}, additionally, we assess a range of variants of the Mamba model, including Mamba, Bi-Mamba and Bi-Mambaformer as Figure \ref{f2}, \ref{f3}.

\vspace{0.3cm} %

\noindent\textbf{Evaluation Metrics.} We adopt HR(Hit Ratio) \cite{alsini2020hit}, NDCG(Normalized Discounted Cumulative Gain) \cite{jeunen2023normalised}, and MRR(Mean Reciprocal Rank) \cite{lu2023optimizing} with truncation at 10 as the evaluation metrics. HR@10, NDCG@10, and MRR@10.  HR is chosen for its straightforward interpretation and ability to measure recall, reflecting how well the system includes relevant items in the top recommendations. NDCG is selected because it considers both the relevance and the order of recommended items, providing a comprehensive assessment of recommendation quality. MRR emphasizes the ranking accuracy of the first relevant item, making it useful for gauging user satisfaction, especially when users tend to look at the top recommendations. Together, these metrics offer a multidimensional evaluation of the recommendation system's performance.

\begin{figure}[h]
    \centering
\begin{minipage}{0.45\textwidth}
\vspace{0.95cm} %
\begin{tikzpicture}
  
  \draw[rounded corners, fill=gray!10] (-0.2,0) rectangle (5.2,6);

  \node at (5,6.2) {x L};

  \node[draw, rounded corners, fill=yellow!30, minimum width=1.5cm, minimum height=0.4cm] (addnorm1) at (2.5,5.5) {Add \& Norm};
  \node[draw, rounded corners, fill=purple!30, minimum width=1.5cm, minimum height=0.4cm] (ffn) at (2.5,4.5) {GLU};

  \node[draw, rounded corners, fill=yellow!30, minimum width=1.5cm, minimum height=0.4cm] (addnorm2) at (1.2,2.9) {Add \& Norm};
  \node[draw, rounded corners, fill=yellow!30, minimum width=1.5cm, minimum height=0.4cm] (addnorm3) at (3.8,2.9) {Add \& Norm};

  \node[draw, rounded corners, fill=pink!80, minimum width=1.5cm, minimum height=0.4cm] (mamba1) at (1.2,1.6) {Mamba};
  \node[draw, rounded corners, fill=pink!80, minimum width=1.5cm, minimum height=0.4cm] (mamba2) at (3.8,1.6) {Mamba};

  \node[draw, circle,fill=white, minimum size=0.4cm] (circle) at (2.5,3.67){  };

  \draw[->] (3.8, 0.82) -- (mamba2);
  \draw[->] (mamba1) -- (addnorm2);
  \draw[->] (ffn) -- (addnorm1);
  \draw[->] (1.2,0.82) -| (mamba1);
  \draw[-] (2.5,-0.2) -- (2.5,0.35); //filter
  \draw[->] (2.5,5.72) -- (2.5,6.5); //output
  \draw[->] (addnorm2) |- (circle);
  \draw[->] (addnorm3) |- (circle);
  \draw[->] (2.5,3.46) -- (2.5,4.28); //circle
  \draw[-] (2.2,3.67) -- (2.8,3.67);
  \draw[dashed, ->] (2.5,0.35) |- (addnorm2);
  \draw[dashed, ->] (2.5,0.35) |- (addnorm3);
  \draw[dashed, -] (2.5,4) -- (3.8,4);
  \draw[dashed, ->] (3.8,4) |- (addnorm1);
  \draw[-] (2.5,0.35) -| (1.2, 0.62);
  \draw[-] (2.5,0.35) -| (3.8, 0.62);
  \draw[->] (mamba2) -- (addnorm3);

  \node at (1.2, 0.75) {\small{\( v_1, v_2, \ldots, v_{n_u} \)}};
  \node at (3.85, 0.75) {\small{\( v_{n_u}, \ldots, v_2, v_1\)}};
  \node at (4.5, 2.2) {\textbf \small{Reverse}};
  \node at (4.5, 0.35) {\textbf \small{Reverse}};
\end{tikzpicture}
\centering
\caption{\textbf{Bi-Mamba4Rec} \cite{liang2024bi}}
\label{f2}
\end{minipage}
\hfill
\begin{minipage}{0.45\textwidth}
\begin{tikzpicture}
  \draw[rounded corners, fill=gray!10] (-0.2,-1) rectangle (5.2,6);

  \node at (5,6.2) {x L};

  \node[draw, rounded corners, fill=red!40, minimum width=3cm, minimum height=0.4cm] (fl) at (2.5,-0.35) {Attention Layer};

  \node[draw, rounded corners, fill=yellow!30, minimum width=1.5cm, minimum height=0.4cm] (addnorm1) at (2.5,5.5) {Add \& Norm};
  \node[draw, rounded corners, fill=purple!30, minimum width=1.5cm, minimum height=0.4cm] (ffn) at (2.5,4.5) {GLU};

  \node[draw, rounded corners, fill=yellow!30, minimum width=1.5cm, minimum height=0.4cm] (addnorm2) at (1.2,2.9) {Add \& Norm};
  \node[draw, rounded corners, fill=yellow!30, minimum width=1.5cm, minimum height=0.4cm] (addnorm3) at (3.8,2.9) {Add \& Norm};

  \node[draw, rounded corners, fill=pink!80, minimum width=1.5cm, minimum height=0.4cm] (mamba1) at (1.2,1.6) {Mamba};
  \node[draw, rounded corners, fill=pink!80, minimum width=1.5cm, minimum height=0.4cm] (mamba2) at (3.8,1.6) {Mamba};

  \node[draw, circle,fill=white, minimum size=0.4cm] (circle) at (2.5,3.67){  };

  \draw[->] (3.8, 0.82) -- (mamba2);
  \draw[->] (mamba1) -- (addnorm2);
  \draw[->] (ffn) -- (addnorm1);
  \draw[->] (1.2,0.82) -| (mamba1);
  \draw[-] (fl) -- (2.5,0.35); //filter
  \draw[->] (2.5,5.72) -- (2.5,6.5); //output
  \draw[->] (addnorm2) |- (circle);
  \draw[->] (addnorm3) |- (circle);
  \draw[->] (2.5,3.46) -- (2.5,4.28); //circle
  \draw[-] (2.2,3.67) -- (2.8,3.67);
  \draw[dashed, ->] (2.5,0.35) |- (addnorm2);
  \draw[dashed, ->] (2.5,0.35) |- (addnorm3);
  \draw[dashed, -] (2.5,4) -- (3.8,4);
  \draw[dashed, ->] (3.8,4) |- (addnorm1);
  \draw[->] (2.5,-1.2) -- (fl);
  \draw[-] (2.5,0.35) -| (1.2, 0.62);
  \draw[-] (2.5,0.35) -| (3.8, 0.62);
  \draw[->] (mamba2) -- (addnorm3);

  \node at (1.2, 0.75) {\small{\( v_1, v_2, \ldots, v_{n_u} \)}};
  \node at (3.85, 0.75) {\small{\( v_{n_u}, \ldots, v_2, v_1\)}};
  \node at (4.5, 2.2) {\textbf \small{Reverse}};
  \node at (4.5, 0.35) {\textbf \small{Reverse}};

\end{tikzpicture}
\centering
\caption{\textbf{Bi-Mambaformer4Rec} \cite{liang2024bi, xu2024integrating} }
\label{f3}
\end{minipage}
\end{figure}

\subsection{Recommendation performance}
Based on the results presented in Table \ref{tab:performance}, the evaluation of models reveals notable trends. RNN-based models, including NARM and GRU4Rec, demonstrate competitive but relatively lower performance compared to attention-based models such as SASRec and BERT4Rec. And the Mamba model variants, outperform both RNN and attention-based models across MovieLens-1M, Amazon-Beauty, and Amazon-Video-Games datasets. Notably, EchoMamba4Rec achieves superior performance compared to baselines based on RNN, Transformer, Mamba and thier hybrid  showcasing its effectiveness in capturing intricate user-item interactions and providing accurate recommendations.

\begin{table*}[h]
\centering
\caption{Recommendation performance. The best results are bold, and the second-best are underlined.}
\label{tab:performance}
\resizebox{\textwidth}{!}{
\begin{tabular}{c|ccc|ccc|ccc}
\toprule
\multicolumn{1}{c}{\multirow{2}{*}{\centering\textbf{Method}}} & \multicolumn{3}{c}{\textbf{MovieLens-1M}} & \multicolumn{3}{c}{\textbf{Amazon-Beauty}} & \multicolumn{3}{c}{\textbf{Amazon-Video-Games}} \\ \cmidrule(lr){2-4} \cmidrule(lr){5-7} \cmidrule(lr){8-10}
& \textbf{HR@10} & \textbf{NDCG@10} & \textbf{MRR@10} & \textbf{HR@10} & \textbf{NDCG@10} & \textbf{MRR@10} & \textbf{HR@10} & \textbf{NDCG@10} & \textbf{MRR@10} \\ [0.5ex]
\hline
NARM   &0.2884 &0.1445 &0.1267 &0.0640 &0.0369 &0.0287 &0.1057 &0.0556 &0.0405 \\
GRU4Rec &0.2642 &0.1464 &0.1106 &0.0664 &0.0365 &0.0274 &0.1101 &0.0573 &0.0414 \\
SASRec &0.2429 &0.1262 &0.0905 &0.0739 &0.0404 &0.0302 &0.1203 &0.0561 &0.0366 \\
BERT4Rec &0.2910 &0.0979 &0.0663 &0.0711 &\underline{0.0426} &0.0309 &0.1106 &0.0508 &0.0324 \\
Mamba4Rec          & 0.3069 & \underline{0.1509} & 0.1223 & 0.0621 & 0.0369 & 0.0292 & 0.1057 & \underline{0.0587} & 0.0441 \\
Bi-Mamba4Rec & 0.3051 & 0.1485 & \underline{0.1398} & \underline{0.0781} & 0.0423 & 0.0313 & 0.1200 & 0.0580 & 0.0392 \\
Bi-Mambaformer4Rec       & \underline{0.3166} & 0.1470 & 0.1106 & 0.0773 & 0.0304 & \textbf{0.0412} & \underline{0.1205} & 0.0451 & \textbf{0.0626} \\
\hline
EchoMamba4Rec & \textbf{0.3215} & \textbf{0.1529} & \textbf{0.1458} & \textbf{0.0833} & \textbf{0.0438} & \underline{0.0318} & \textbf{0.1254} & \textbf{0.0633} & \underline{0.0445} \\
\bottomrule
\end{tabular}}
\end{table*}

\subsection{Efficiency performance}
Table \ref{tab:3} shows the efficiency performance of attention-based models, Mamba4Rec, and EchoMamba4Rec in detail. All measurements were conducted on an RTX 4090 GPU using the Amazon-Video Games dataset. EchoMamba4Rec exhibits superior performance in terms of training time per epoch, while the GPU memory cost is close to that of Mamba. These results demonstrate that EchoMamba4Rec achieves better efficiency when controlling memory usage compared to attention-based models and Mamba.

\vspace{0.3cm} %

\begin{table}[h!]
\centering
\caption{Efficiency performance on amazon-video games}
\label{tab:3}
\begin{tabular}{c|c|c|c}
\toprule
\textbf{Method} & \textbf{GPU memory (GB)} & \textbf{Training time (s)} & \textbf{Inference time (s)}
\\
\midrule
SASRec & 6.58 & 163.543 & 0.3015\\
BERT4Rec & 6.68 & 215.800 & 0.3218\\
Mamba4Rec & \textbf{1.41} & \underline{98.759} 
& \underline{0.0305}
\\
\midrule
EchoMamba4Rec & \underline{1.61}  & \textbf{70.187} & \textbf{0.0252}\\
\bottomrule
\end{tabular}

\end{table}

\section{Future work}\label{sec5}

In future work, we plan to explore the following directions to further enhance the capabilities and applicability of the EchoMamba4Rec model:

\vspace{0.3cm}

\noindent \textbf{Advanced Filtering Techniques:} investigate the integration of additional advanced filtering methods, such as wavelet transforms \cite{daubechies1992ten} and adaptive filtering \cite{haykin2002adaptive}, to improve the model's ability to capture complex and nuanced patterns in user behavior sequences.
    
\noindent \textbf{Cross-Domain Evaluation:} test the EchoMamba4Rec model on a wider range of datasets from different domains, such as e-commerce, streaming services, and social media \cite{zhang2021group}, to evaluate its generalizability and robustness across various application scenarios.

\sloppy
\noindent \textbf{Real-Time Implementation:} explore the feasibility of implementing EchoMamba4Rec in real-time recommendation systems to assess its performance in a live environment \cite{hong2008routledge}, focusing on response time and scalability.
\fussy

\noindent \textbf{User Privacy and Fairness:} investigate the implications of user privacy and fairness in the recommendations generated by EchoMamba4Rec, ensuring that the model adheres to ethical guidelines and provides equitable recommendations for all users \cite{kamishima2012enhancement}.

\section{Acknowledgement}\label{sec6}
    
The authors would like to appreciate the support from
the Interdisciplinary Intelligence Super Computer Cen-
ter of Beijing Normal University at Zhuhai.
This work was funded by the Natural Science Foundation of China (12271047); Guangdong Provincial Key Laboratory of Interdisciplinary Research and Application for Data Science, BNU-HKBU United International College (2022B1212010006); UIC research grant (R0400001-22; UICR0400008-21; UICR0400036-21CTL; UICR04202405-21); Guangdong College Enhancement and Innovation Program (2021ZDZX1046).


\bibliography{sn-bibliography}
\begin{appendices}

\section{Definition}\label{secA1}

\subsection{Hit Ratio (HR)}
Hit Ratio measures the fraction of times the correct item appears in the top \(N\) recommended items for each user. It can be formally defined as:
\[HR = \frac{1}{|U|} \sum_{u \in U} \mathbb{I}(\text{rank}(u) \leq N).\]

\noindent\( U \) is the set of all users. \( \text{rank}(u) \) is the rank position of the correct item for user \( u \) in the list of recommendations. \( \mathbb{I}(\cdot) \) is the indicator function, which returns 1 if the condition is true and 0 otherwise.

\subsection{Normalized Discounted Cumulative Gain (NDCG)}
Normalized Discounted Cumulative Gain assesses the ranking quality of the recommended list by considering the position of the relevant items. It is defined as:
\[NDCG = \frac{1}{|U|} \sum_{u \in U} \frac{DCG(u)}{IDCG(u)}.\]

\noindent \( DCG(u) \) is the Discounted Cumulative Gain for user \( u \), calculated as:
 \[  DCG(u) = \sum_{i=1}^{N} \frac{2^{rel_i(u)} - 1}{\log_2(i+1)}.\]
 \( IDCG(u) \) is the Ideal Discounted Cumulative Gain for user \( u \), which is the maximum possible DCG that can be achieved, calculated by ordering the true relevant items in the optimal ranking.
\( rel_i(u) \) is the relevance score of the item at position \( i \) for user \( u \).

\subsection{Mean Reciprocal Rank (MRR)}
Mean Reciprocal Rank evaluates the average of the reciprocal ranks of the first relevant item in the recommended list. It is defined as:
\[MRR = \frac{1}{|U|} \sum_{u \in U} \frac{1}{\text{rank}(u)}.\]

\noindent \( \text{rank}(u) \) is the rank position of the first relevant item for user \( u \) in the list of recommendations.




\end{appendices}

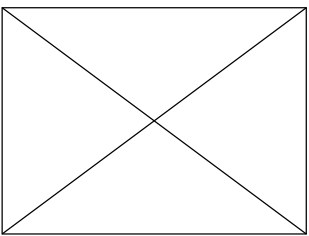
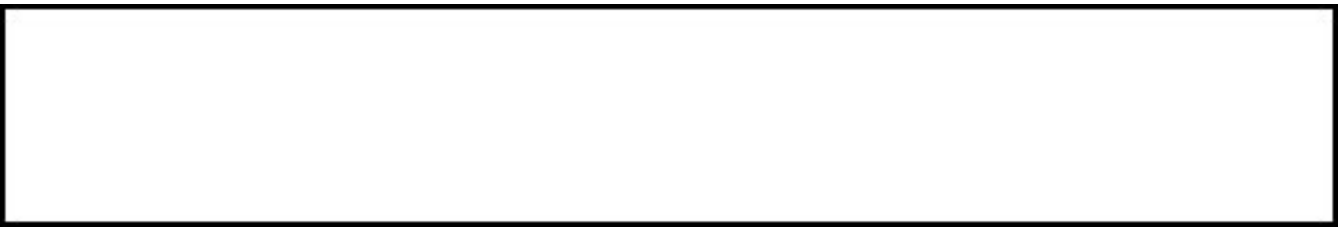
\end{document}